# Aquila-plus: Prompt-Driven Visual-Language Models for Pixel-Level Remote Sensing Image Understanding

Kaixuan Lu

Aerospace Information Research Institute, Chinese Academy of Sciences, Beijing 100094, China

**Abstract**

The recent development of vision language models (VLMs) has led to significant advances in visual-language integration through visual instruction tuning, and they have rapidly evolved in the field of remote sensing image understanding, demonstrating their powerful capabilities. However, existing RSVLMs mainly focus on image-level or frame-level understanding, making it difficult to achieve fine-grained pixel-level visual-language alignment. Additionally, the lack of mask-based instructional data limits their further development. In this paper, we propose a mask-text instruction tuning method called Aquila-plus, which extends the capabilities of RSVLMs to achieve pixel-level visual understanding by incorporating fine-grained mask regions into language instructions. To achieve this, we first meticulously constructed a mask region-text dataset containing 100K samples, and then designed a visual-language model by injecting pixel-level representations into a large language model (LLM). Specifically, Aquila-plus uses a convolutional

CLIP as the visual encoder and employs a mask-aware visual extractor to extract precise visual mask features from high-resolution inputs. Experimental results demonstrate that Aquila-plus outperforms existing methods in various region understanding tasks, showcasing its novel capabilities in pixel-level instruction tuning.

## 1. Instruction

Vision language models (VLMs) are crucial building blocks for creating a general-purpose visual assistant, and they have gained increasing popularity in the research community in recent years. Although recent VLMs such as LLaVA, MiniGPT-4, Otter, Instruct-BLIP, Qwen-VL, and LLaVA-1.5 have demonstrated remarkable achievements in instruction following and visual reasoning capabilities, most of them perform visual-language alignment at the image level, using image-text pairs. The lack of region-level alignment makes it challenging for these models to handle fine-grained image understanding tasks, such as region classification, description, and reasoning.

Compared to coarse bounding boxes, using fine-grained masks as reference inputs allows for a more precise representation of objects. By training with numerous high-quality masks, the recently developed SAM supports using simple bounding boxes or points as prompts, and it excels in zero-shot segmentation quality for objects, parts, or subparts. Several studies, such as HQ-SAM, have further enhanced SAM's capabilities in fine-grained segmentation and generalization, making segmentation more practical in real-world applications. However, these models cannot provide primary semantic labels, let alone detailed semantic attributes and descriptions. As a result, existing methods are limited when it comes to understanding real-world scenes that contain fine-grained multimodal information.

This paper presents a novel method, Aquila-plus, aimed at expanding the capability of VLMs in fine-grained pixel-level understanding. To achieve this, we propose a mask-aware visual extractor to capture precise visual mask features at different granularities. These visual features are interleaved with language instructions, forming an input sequence for the LLM. To facilitate the use of high-resolution inputs, we employ convolutional CLIP as the visual encoder. Compared to ViT-based models, convolutional CLIP performs well at higher input resolutions and offers greater efficiency and robustness. With this design, Aquila-plus can achieve fine semantic understanding of object-level and part-level regions, providing primary object categories, detailed object attributes, and more complex scene descriptions.

To achieve fine-grained pixel-level alignment between visual and language features, we meticulously constructed a large-scale mask-based region-text dataset called Aquila-plus-100K, in which each region's mask and text description are carefully annotated. Most of the data in this dataset come from publicly available datasets and are formatted in an instruction-following style using well-designed prompt templates, covering both object-level and part-level samples. The dataset includes not only detailed descriptions and dialogues but also enriched attribute information. Additionally, we enhanced Aquila-plus's response robustness and flexibility by introducing spatial-aware and category-aware negative sample mining as well as short-form response instructions.

Through visual instruction tuning, our proposed model achieves new capabilities that go beyond frame-level and image-level understanding. As shown in Figure 1-(b), Aquila-plus can generate fine-grained semantics based on class-agnostic masks from the existing SAM. Extensive experimental results demonstrate that our approach outperforms others in open-vocabulary recognition, object classification, description and reasoning, as well as object hallucination tasks. The main contributions of this paper are summarized as follows:

- We propose a novel method, Aquila-plus, which enables multimodal large language models (VLMs) to achieve pixel-level instruction tuning, thereby facilitating fine-grained and open-world visual understanding.

- We constructed a large-scale instruction tuning dataset for the remote

sensing domain, named Aquila-plus-100K, containing object-level and part-level mask-text pairs to enhance the model's robustness and flexibility.

- Our method, as a fine visual understanding approach, outperforms the current state-of-the-art methods across various region understanding tasks.

## 2. Related Work

### 2.1 Vision Language Models

Large language models (LLMs), such as GPT-3, Flan-T5, PaLM, and LLaMA, have achieved significant progress in natural language processing (NLP) research. These advancements have driven the development of multimodal language models, enabling applications like ChatGPT by expanding training data and increasing model scale. The success of LLMs and VLMs has also inspired research in computer vision, making multimodal context learning possible. Recent research has increasingly focused on leveraging pretrained LLMs for visual instruction tuning. Models such as LLaVA, MiniGPT-4, mPLUG-Owl, Otter, Instruct-BLIP, Qwen-VL, and LLaVA-1.5 have demonstrated impressive capabilities in instruction following and visual reasoning tasks. However, these models are limited in multimodal tasks at the image level and struggle when it comes to referring to specific regions for finer understanding.

## 2.2 Region-Level Image Understanding

In the context of region-level image understanding, the process begins with locating regions of potential interest, followed by deeper visual comprehension. The Segment Anything Model (SAM), through training on numerous high-quality masks, has shown outstanding performance in zero-shot object, part, and sub-part segmentation tasks. Since the original SAM does not provide semantic labels, several methods, such as SEEM, HIPIE, and Semantic SAM, have extended the model to predict semantic categories for the masks. However, basic semantic labels are often insufficient for real-world applications. Thus, there is a need to introduce additional semantic information, such as color, position, or even comprehensive descriptions for scene understanding and reasoning. Although some works have achieved pixel-level alignment, they fail to provide region-based descriptions.

Recent studies, such as GPT4RoI, PVIT, Kosmos-2, Shikra, Ferret, and GLaMM, have enabled MLLMs to achieve region-level image understanding. However, most of these approaches use bounding boxes as reference input regions, which may introduce irrelevant background features, resulting in less precise region-text alignment during visual instruction tuning. Furthermore, these models only allow for smaller input image sizes, such as 224×224, which makes it difficult to analyze details in densely populated object regions. To address these issues, this paper introduces a pixel-level understanding method based on LLMs. Our method supports using input masks for region references and allows for higher image resolutions. Additionally, we constructed a dataset containing mask-text pairs to facilitate instruction-based learning.

## 3. Aquila-plus-100K Dataset

In this section, we introduce Aquila-plus-100K, an instruction-based remote sensing dataset containing mask-text pairs, consisting of approximately 100K multimodal dialogues aimed at encouraging VLMs to perform pixel-level image understanding. Specifically, Aquila-plus-100K comprises mask-text instruction data at both object-level and part-level, created based on publicly available datasets. To ensure that the data meets the requirements for instruction-following tasks, we utilized GPT-4 with carefully designed prompt templates to generate high-quality mask-text pairs. Furthermore, to enhance the robustness and flexibility of responses, we introduced negative sample mining techniques and short-form response prompts. Figure 1 illustrates an example from the Aquila-plus-100K dataset.

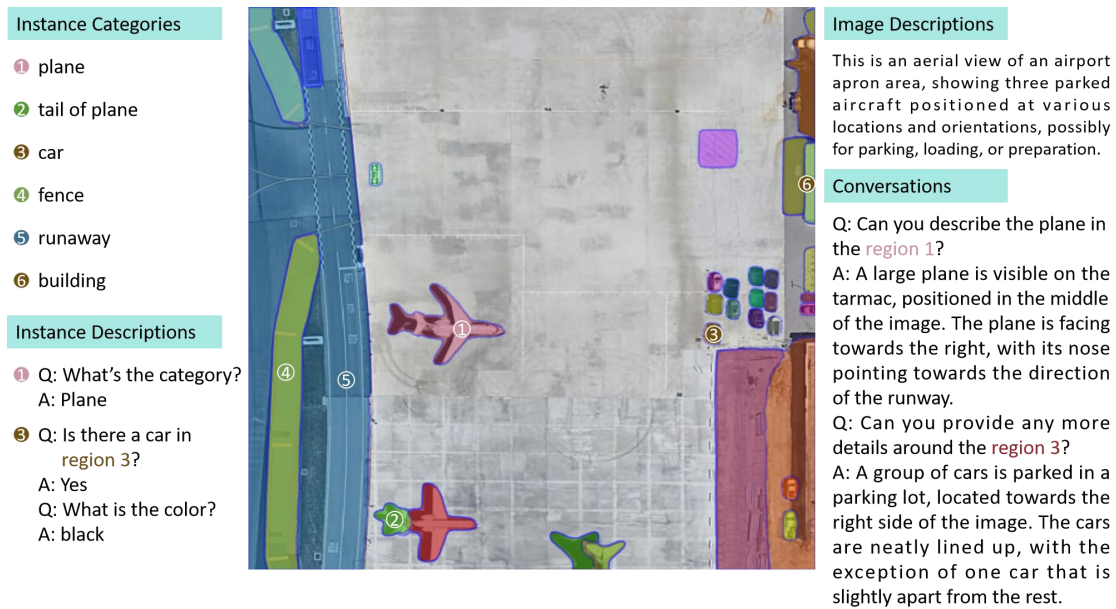

Figure 1 An example of the Aquila-plus-100K dataset.

## 4. Aquila-plus Method

### 4.1 Model Architecture

An overview of the Aquila-plus architecture is illustrated in Figure 2. Aquila-plus consists of an image-level visual encoder, a pixel-level mask-aware visual extractor, and a large language model (LLM). Given an image, a reference mask region, and an input language prompt, we perform tokenization and transformation to obtain embedded representations. The interleaved sequence of mask features and language embeddings is then fed into the LLM to achieve fine semantic understanding.

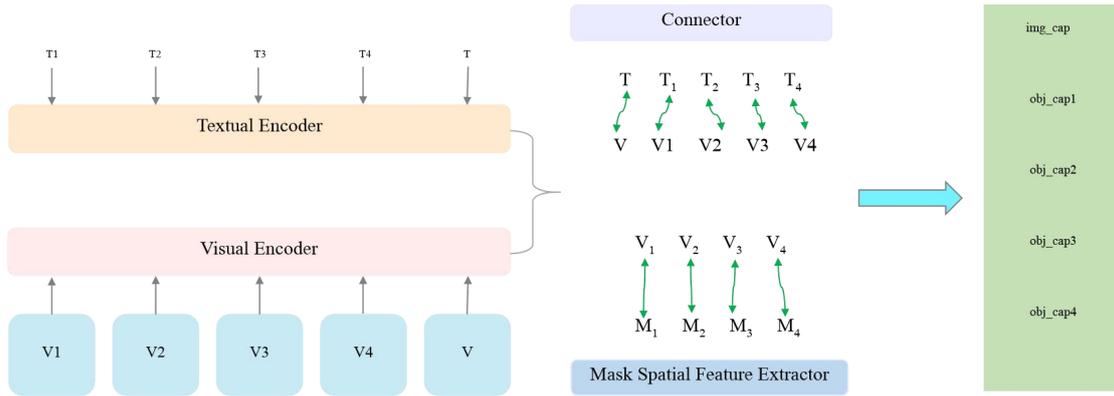

Figure 2 Overview of the Aquila-plus architecture.

## 4.2 Convolutional CLIP Visual Encoder

Most visual encoders in VLMs utilize ViT-based CLIP models, such as those used in LLaVA. However, lower input resolutions make it challenging to achieve fine-grained image understanding for pixel-level representations, especially in small regions. To address this issue, we introduce convolutional CLIP models, such as ResNet and ConvNeXt, as the visual encoder. The CNN-based convolutional CLIP model performs well across various input resolutions and offers higher generalization ability compared to ViT-based CLIP models. Moreover, the multi-scale feature maps generated by CNN can be directly used for subsequent feature extraction of each object region. In implementation, we choose the ConvNeXt-Large CLIP model as the visual encoder and use the output from the last stage as the image-level feature.

## 4.3 Mask Spatial Feature Extractor

Unlike previous region-level approaches that used sparse bounding boxes as reference inputs (e.g., PVIT, Shikra), Aquila-plus utilizes fine-grained mask region representations. To capture pixel-level features of each object region, we propose a mask-aware visual extractor that not only encodes mask-level visual features but also gathers spatial positional information for each region.

Specifically, we first apply a mask pooling operation to the multi-layer image features produced by the visual encoder, pooling all the features that fall within the mask region. We then pass each layer of features through a linear projection layer to generate region-level embeddings of the same dimension, and we fuse multi-layer features by summation. To adapt and produce visual mask tokens, we further use a multi-layer perceptron (MLP) layer for processing.

To retain the spatial geometric relationship of object regions, we encode pixel-level spatial relations using the binary mask of each object region. First, we resize the mask to a size of 224×224, then flatten and project it to generate spatial tokens. Finally, we combine the visual mask tokens with their corresponding spatial tokens to form the embedding representation for each mask region.

### 4.4 Tokenization

As shown in Figure 2, we input the image into the pretrained visual encoder, ConvNeXt-Large CLIP, to extract image-level embedding representations. For textual information, we tokenize the text sequence and project it to obtain text embeddings. For mask-based regions, we define a special token as a placeholder <region> and replace it with mask tokens and spatial tokens, such as <mask> and <position>. When referring to an object region in the text input, <region> is appended after the region name, such as "region1" or "region2." In this way, the mask regions blend seamlessly with the text to form complete sentences, which are subsequently tokenized together.

In addition to the user's instruction, we introduce a prefix prompt: "<image>\n An overview of the given image is provided." Here, <image> is a special placeholder token that is replaced by the image-level embeddings from the visual encoder. All image-level and region-level visual tokens are interleaved with text tokens and input into the LLM for understanding the image and responding to the user's instructions regarding different object regions. We use the Vicuna model as the LLM, which is an instruction-tuned decoder version based on LLaMA.

### 4.5 Training

The training process for the Aquila-plus model consists of three stages, with supervision in all stages achieved by minimizing the next token prediction loss.

Stage 1: Image-Text Alignment Pretraining

Using the ConvNeXt-Large convolutional CLIP visual encoder, we begin by training image-level features and a language connector to achieve image-text feature alignment. In this stage, Aquila-plus consists of a pretrained visual encoder, a pretrained LLM, an image-level projector and a region-level projector. We employ a multi-layer perceptron (MLP) as the visual-language connector to enhance the model's multimodal capabilities. During this stage, only the image-level projector and the region-level are trained, while the visual encoder and LLM remain frozen.

Stage 2: End-to-End Fine-Tuning

In this stage, we fix the weights of the visual encoder and fine-tune the image-level projector, region-level projector, and LLM of Aquila-plus. The focus is on expanding Aquila-plus's capabilities to accurately follow user instructions and handle complex pixel-level region understanding tasks. During this stage, we use the constructed Aquila-plus-100K dataset.

## 5. Experiments

### 5.1 Experimental Results

To evaluate the effectiveness of our proposed model, Aquila-plus, we conducted multiple experiments to demonstrate its capabilities in tasks such as pixel-level region recognition, classification, and complex description and reasoning, as shown in the Figure 3.

| | | |
|---|---|---|
| 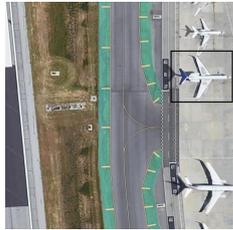 | 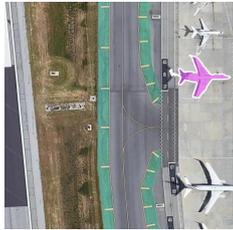 | A large airplane is situated on the right side of the image, near the edge. It's on the runway, ready for takeoff. The plane is mostly white, with a distinct blue tail. |
| 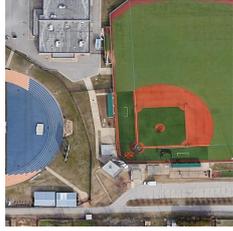 | 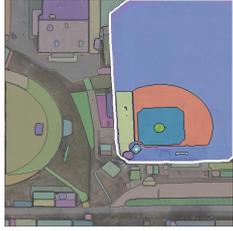 | A large, green baseball field is depicted, which is currently being used for a game. The field is well-maintained and has a clearly visible diamond, indicating the placement of the baseball game's center. |
| 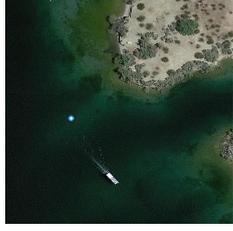 | 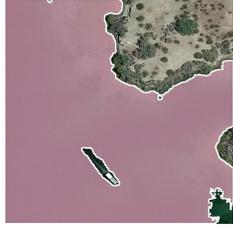 | The image shows a large portion of a green lake, covering most of the area. The water is calm and the lake appears to be quite vast. |
| 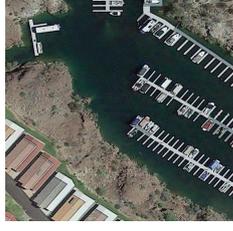 | 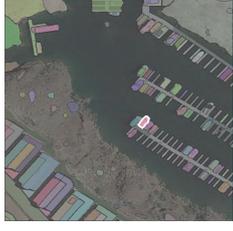 | A large boat is visible in the water, taking up a significant portion of the view. It's distinguishable by its blue hull and a white box or structure on it. The boat appears to be docked, and it's the second such boat from the left in the image. |
| 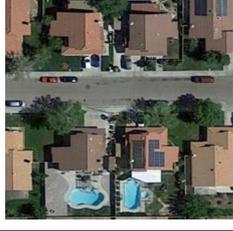 | 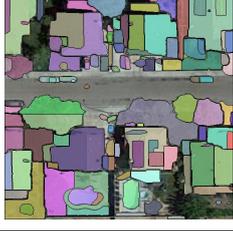 | A pool is located in the middle of the scene, surrounded by two other pools. This pool is not the closest nor the furthest from the viewer's perspective. It is the middle pool in the row of three. |
| 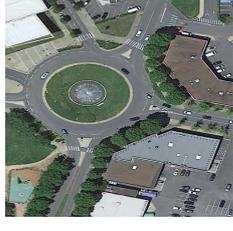 | 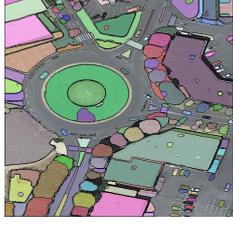 | A round, grassy area is situated in the middle of the scene, surrounded by a white circle. This area, which could be a park or a traffic circle, is positioned between two roads. It's a significant feature in the landscape, attracting attention with its lush greenery and central location. |

Figure 3. Examples of feeding Aquila-plus with class-agnostic masks from off-the shelf SAM.

## 6. Conclusion

This paper presents a novel method, Aquila-plus, that introduces pixel-level mask region referencing into language instructions, significantly enhancing the fine-grained visual understanding capabilities of VLMs. By incorporating a mask-aware visual extractor and employing a convolutional CLIP model as the visual encoder, Aquila-plus achieves region-based image understanding capabilities. To achieve fine-grained pixel-level alignment between vision and language, we meticulously constructed a dataset named Aquila-plus-100K, which contains 724K mask-text pairs. Aquila-plus demonstrates superior performance across various region understanding tasks, achieving new state-of-the-art results. We hope that the Aquila-plus-100K dataset and the Aquila-plus model will advance visual region understanding in practical applications of VLMs.

## 7. Reference

[1]Zheng Z, Zhong Y, Ma A, et al. HyNet: Hyper-scale object detection network framework for multiple spatial resolution remote sensing imagery[J]. ISPRS Journal of Photogrammetry and Remote Sensing,

2020, 166: 1-14.

[2]Chen H, Yang W, Liu L, et al. Coarse-to-fine semantic segmentation of satellite images[J]. ISPRS Journal of Photogrammetry and Remote Sensing, 2024, 217: 1-17.

[3]Ning X, Zhang H, Zhang R, et al. Multi-stage progressive change detection on high resolution remote sensing imagery[J]. ISPRS Journal of Photogrammetry and Remote Sensing, 2024, 207: 231-244.

[4]OpenAI. Chatgpt. https://openai.com/blog/chatgpt/, 2023. 1, 2, 3, 4, 6, 7, 11, 13, 15, 16.

[5]OpenAI. Gpt-4 technical report, 2023. 1, 5, 6, 15.

[6]Wei-Lin Chiang, Zhuohan Li, Zi Lin, Ying Sheng, Zhanghao Wu, Hao Zhang, Lianmin Zheng, Siyuan Zhuang, Yonghao Zhuang, Joseph E. Gonzalez, Ion Stoica, and Eric P. Xing. Vicuna: An open-source chatbot impressing gpt-4 with 90%* chatgpt quality. https://lmsys.org/blog/2023-03-30-vicuna/, 2023. 2, 3, 4, 7, 8, 9, 17.

[7]Radford A, Kim J W, Hallacy C, et al. Learning transferable visual models from natural language supervision[C]//International conference on machine learning. PMLR, 2021: 8748-8763.

[8]Li Y, Zhang Y, Wang C, et al. Mini-gemini: Mining the potential of multi-modality vision language models[J]. arXiv preprint arXiv:2403.18814, 2024.

[9]Peng Z, Wang W, Dong L, et al. Kosmos-2: Grounding multimodal

large language models to the world[J]. arXiv preprint arXiv:2306.14824, 2023.

[10]Zhang Z, Zhao T, Guo Y, et al. Rs5m: A large scale vision-language dataset for remote sensing vision-language foundation model[J]. arXiv preprint arXiv:2306.11300, 2023.

[11]Wang Z, Prabha R, Huang T, et al. Skyscript: A large and semantically diverse vision-language dataset for remote sensing[C]//Proceedings of the AAAI Conference on Artificial Intelligence. 2024, 38(6): 5805-5813.

[12]Zhang W, Cai M, Zhang T, et al. Earthgpt: A universal multi-modal large language model for multi-sensor image comprehension in remote sensing domain[J]. IEEE Transactions on Geoscience and Remote Sensing, 2024.

[13]Hu Y, Yuan J, Wen C, et al. Rsgpt: A remote sensing vision language model and benchmark[J]. arXiv preprint arXiv:2307.15266, 2023.

[14]Vaswani A. Attention is all you need[J]. Advances in Neural Information Processing Systems, 2017.

[15]Liu H, Li C, Li Y, et al. Improved baselines with visual instruction tuning[C]//Proceedings of the IEEE/CVF Conference on Computer Vision and Pattern Recognition. 2024: 26296-26306.

[16]Danny Driess, Fei Xia, Mehdi SM Sajjadi, Corey Lynch, Aakanksha Chowdhery, Brian Ichter, Ayzaan Wahid, Jonathan Tompson, Quan Vuong,

Tianhe Yu, et al. PaLM-E: An embodied multimodal language model. arXiv preprint arXiv:2303.03378, 2023. 2

[17]Li J, Li D, Xiong C, et al. Blip: Bootstrapping language-image pre-training for unified vision-language understanding and generation[C]//International conference on machine learning. PMLR, 2022: 12888-12900.

[18]Bai J, Bai S, Yang S, et al. Qwen-vl: A frontier large vision-language model with versatile abilities[J]. arXiv preprint arXiv:2308.12966, 2023.

[19]Li Z, Yang B, Liu Q, et al. Monkey: Image resolution and text label are important things for large multi-modal models[C]//Proceedings of the IEEE/CVF Conference on Computer Vision and Pattern Recognition. 2024: 26763-26773.

[20]Ye Q, Xu H, Xu G, et al. mplug-owl: Modularization empowers large language models with multimodality[J]. arXiv preprint arXiv:2304.14178, 2023.

[21]Luo J, Pang Z, Zhang Y, et al. Skysensegpt: A fine-grained instruction tuning dataset and model for remote sensing vision-language understanding[J]. arXiv preprint arXiv:2406.10100, 2024.

[22]Liu Z, Mao H, Wu C Y, et al. A convnet for the 2020s[C]//Proceedings of the IEEE/CVF conference on computer vision and pattern recognition. 2022: 11976-11986.

[23]Dosovitskiy A, Beyer L, Kolesnikov A, et al. An image is worth

16x16 words: Transformers for image recognition at scale[J]. arXiv preprint arXiv:2010.11929, 2020.

[24]Devlin J. Bert: Pre-training of deep bidirectional transformers for language understanding[J]. arXiv preprint arXiv:1810.04805, 2018.

[25]Touvron H, Lavril T, Izacard G, et al. Llama: Open and efficient foundation language models[J]. arXiv preprint arXiv:2302.13971, 2023.

[26]Touvron H, Martin L, Stone K, et al. Llama 2: Open foundation and fine-tuned chat models[J]. arXiv preprint arXiv:2307.09288, 2023.

[27]Team G, Mesnard T, Hardin C, et al. Gemma: Open models based on gemini research and technology[J]. arXiv preprint arXiv:2403.08295, 2024.

[28]Radford A. Improving language understanding by generative pre-training[J]. 2018.

[29]Brown T B. Language models are few-shot learners[J]. arXiv preprint arXiv:2005.14165, 2020.

[30]Jiang A Q, Sablayrolles A, Mensch A, et al. Mistral 7B[J]. arXiv preprint arXiv:2310.06825, 2023.

[31]Zhu D, Chen J, Shen X, et al. Minigpt-4: Enhancing vision-language understanding with advanced large language models[J]. arXiv preprint arXiv:2304.10592, 2023.

[32]Yang Z, Li L, Wang J, et al. Mm-react: Prompting chatgpt for multimodal reasoning and action[J]. arXiv preprint arXiv:2303.11381,

2023.

[33]Wu C, Yin S, Qi W, et al. Visual chatgpt: Talking, drawing and editing with visual foundation models[J]. arXiv preprint arXiv:2303.04671, 2023.

[34]Yang R, Song L, Li Y, et al. Gpt4tools: Teaching large language model to use tools via self-instruction[J]. Advances in Neural Information Processing Systems, 2024, 36.

[35]Hu E J, Shen Y, Wallis P, et al. Lora: Low-rank adaptation of large language models[J]. arXiv preprint arXiv:2106.09685, 2021.

[36]Alayrac J B, Donahue J, Luc P, et al. Flamingo: a visual language model for few-shot learning[J]. Advances in neural information processing systems, 2022, 35: 23716-23736.

[37]Li J, Li D, Savarese S, et al. Blip-2: Bootstrapping language-image pre-training with frozen image encoders and large language models[C]//International conference on machine learning. PMLR, 2023: 19730-19742.

[38]Liu H, Li C, Wu Q, et al. Visual instruction tuning[J]. Advances in neural information processing systems, 2024, 36.

[39]Liu H, Li C, Li Y, et al. Improved baselines with visual instruction tuning[C]//Proceedings of the IEEE/CVF Conference on Computer Vision and Pattern Recognition. 2024: 26296-26306.

[40]Liu S, Cheng H, Liu H, et al. Llava-plus: Learning to use tools for

creating multimodal agents[J]. arXiv preprint arXiv:2311.05437, 2023.

[41]Chen K, Zhang Z, Zeng W, et al. Shikra: Unleashing multimodal llm's referential dialogue magic[J]. arXiv preprint arXiv:2306.15195, 2023.

[42]Zhou Y, Feng L, Ke Y, et al. Towards Vision-Language Geo-Foundation Model: A Survey[J]. arXiv preprint arXiv:2406.09385, 2024.

[43]Liu F, Chen D, Guan Z, et al. Remoteclip: A vision language foundation model for remote sensing[J]. IEEE Transactions on Geoscience and Remote Sensing, 2024.

[44]Zhang Z, Zhao T, Guo Y, et al. RS5M and GeoRSCLIP: A large scale vision-language dataset and a large vision-language model for remote sensing[J]. IEEE Transactions on Geoscience and Remote Sensing, 2024.

[45]Kuckreja K, Danish M S, Naseer M, et al. Geochat: Grounded large vision-language model for remote sensing[C]//Proceedings of the IEEE/CVF Conference on Computer Vision and Pattern Recognition. 2024: 27831-27840.

[46]Zhan Y, Xiong Z, Yuan Y. Skyeyegpt: Unifying remote sensing vision-language tasks via instruction tuning with large language model[J]. arXiv preprint arXiv:2401.09712, 2024.

[47]Luo J, Pang Z, Zhang Y, et al. Skysensegpt: A fine-grained instruction tuning dataset and model for remote sensing vision-language

understanding[J]. arXiv preprint arXiv:2406.10100, 2024.

[48]Pang C, Wu J, Li J, et al. H2RSVLM: Towards Helpful and Honest Remote Sensing Large Vision Language Model[J]. arXiv preprint arXiv:2403.20213, 2024.

[49]Muhtar D, Li Z, Gu F, et al. Lhrs-bot: Empowering remote sensing with vgi-enhanced large multimodal language model[J]. arXiv preprint arXiv:2402.02544, 2024.

[50]Dubey A, Jauhri A, Pandey A, et al. The llama 3 herd of models[J]. arXiv preprint arXiv:2407.21783, 2024.

[51]Wang W, Lv Q, Yu W, et al. Cogvlm: Visual expert for pretrained language models[J]. arXiv preprint arXiv:2311.03079, 2023.

[52]Bashmal L, Bazi Y, Al Rahhal M M, et al. Capera: Captioning events in aerial videos[J]. Remote Sensing, 2023, 15(8): 2139.

[53]Qu B, Li X, Tao D, et al. Deep semantic understanding of high resolution remote sensing image[C]//2016 International conference on computer, information and telecommunication systems (Cits). IEEE, 2016: 1-5.

[54]Cheng Q, Huang H, Xu Y, et al. NWPU-captions dataset and MLCA-net for remote sensing image captioning[J]. IEEE Transactions on Geoscience and Remote Sensing, 2022, 60: 1-19.

[55]Lu X, Wang B, Zheng X, et al. Exploring models and data for remote sensing image caption generation[J]. IEEE Transactions on Geoscience

and Remote Sensing, 2017, 56(4): 2183-2195.

[56]Yuan Z, Zhang W, Fu K, et al. Exploring a fine-grained multiscale method for cross-modal remote sensing image retrieval[J]. arXiv preprint arXiv:2204.09868, 2022.

[57]Lobry S, Marcos D, Murray J, et al. RSVQA: Visual question answering for remote sensing data[J]. IEEE Transactions on Geoscience and Remote Sensing, 2020, 58(12): 8555-8566.

[58]Xia G S, Yang W, Delon J, et al. Structural high-resolution satellite image indexing[C]//ISPRS TC VII Symposium-100 Years ISPRS. 2010, 38: 298-303.

[59]Papineni K, Roukos S, Ward T, et al. Bleu: a method for automatic evaluation of machine translation[C]//Proceedings of the 40th annual meeting of the Association for Computational Linguistics. 2002: 311-318.